\documentclass[conference]{IEEEtran}
\IEEEoverridecommandlockouts

\usepackage[american]{babel}
\usepackage{float}
\usepackage[bookmarks=false]{hyperref}

\usepackage{cite}
\usepackage{amsmath,amssymb,amsfonts}
\usepackage{algorithmicx}
\usepackage{graphicx}
\usepackage{textcomp}
\usepackage{xcolor}
\usepackage{numprint}
\usepackage{siunitx}

\usepackage[utf8]{inputenc} 
\usepackage[T1]{fontenc}    
\usepackage{url}            
\usepackage{booktabs}       
\usepackage{amsfonts}       
\usepackage{nicefrac}       
\usepackage{microtype}      

\usepackage{amsmath}
\usepackage{todonotes}
\usepackage{natbib}

\usepackage[linesnumbered,ruled,vlined]{algorithm2e}
\usepackage{dsfont}

\hypersetup{backref=true,       
    pagebackref=true,               
    hyperindex=true,                
    colorlinks=true,                
    breaklinks=true,                
    urlcolor= black,                
    linkcolor= blue,                
    bookmarks=true,                 
    bookmarksopen=false,
    filecolor=black,
    citecolor=blue,
    linkbordercolor=blue
}

\newcommand{\fs}{Feedback Signal\xspace}

\def\BibTeX{{\rm B\kern-.05em{\sc i\kern-.025em b}\kern-.08em
    T\kern-.1667em\lower.7ex\hbox{E}\kern-.125emX}}
\begin{document}

\title{I'm Sorry Dave, I'm Afraid I Can't Do That \newline Deep $Q$-Learning from Forbidden Actions
}

\author{
\IEEEauthorblockN{Mathieu Seurin\footnote{contact author}}
\IEEEauthorblockA{Univ. Lille, CNRS, Inria\\ UMR 9189 CRIStAL \\
mathieu.seurin@inria.fr}
\and
\IEEEauthorblockN{Philippe Preux}
\IEEEauthorblockA{Univ. Lille, CNRS, Inria\\ UMR 9189 CRIStAL \\
philippe.preux@inria.fr}
\and
\IEEEauthorblockN{Olivier Pietquin}
\IEEEauthorblockA{Google Research \\ Brain Team \\
pietquin@google.com
}
}

\maketitle

\begin{abstract}
The use of Reinforcement Learning (RL) is still restricted to simulation or to enhance human-operated systems through recommendations.
Real-world environments (\textit{e.g.} industrial robots or power grids) are generally designed with safety constraints in mind implemented in the shape of valid actions masks or contingency controllers. For example, the range of motion and the angles of the motors of a robot can be limited to physical boundaries. Violating constraints thus results in rejected actions or entering in a safe mode driven by an external controller, making RL agents incapable of learning from their mistakes. 
In this paper, we propose a simple modification of a state-of-the-art deep RL algorithm (DQN), enabling learning from forbidden actions. To do so, the standard $Q$-learning update is enhanced with an extra safety loss inspired by structured classification. We empirically show that it reduces the number of hit constraints during the learning phase and accelerates convergence to near-optimal policies compared to using standard DQN. Experiments are done on a Visual Grid World Environment and the TextWorld domain.
\end{abstract}


\section{Introduction}

Reinforcement Learning (RL)~\citep{sutton2018reinforcement} is the main machine learning answer to address sequential decision-making problems under uncertainty.
Its genericity allows application to a variety of domains such as Robotics \citep{levine2016end, gu2017deep}, Resource Management \citep{mao2016resource}, Chemical reaction \citep{zhou2017optimizing}, Traffic-light Management \citep{el2013multiagent}, Spoken Dialogue Systems \citep{lemon2012data} and sparked interest in other industrial applications. Bringing reinforcement learning to critical systems such as Surgery \citep{hashimoto2018artificial}, Dam Management \cite{wang2012dam} or Autonomous Driving \citep{sallab2017deep, leurent2019social} remains one of the most interesting open challenges in machine learning. Unexpected behavior, inability to handle uncertainty and low sampling efficiency are the Achilles heels of real-world RL \citep{dulac2019challenges}. 

The major issue is that designing RL agents able to measure uncertainty about their internal estimates (\textit{e.g.} their state, the outcome of their actions) is difficult \citep{geist2011managing} and especially when associated with recent Deep Learning methods \citep{o2018uncertainty}. Even if available, using a measure of uncertainty for safety is not straightforward either \citep{daubigney2011uncertainty,bellemare2017distributional}. Uncertainty is especially harmful when agents are meant to control a physical system, where a wrong action can lead to catastrophic consequences (\textit{e.g.} damaging material or endangering people). 

Fortunately, many real-world systems are equipped with contingency measures, in the form of \textit{forbidden actions} or external controller taking over when the system is misbehaving. For example, over-temperature monitoring regulates servo-motors present in robots, preventing the motors from reaching their heat limit. Cleaning robots also automatically u-turn in the presence of an obstacle, preventing any damage to the machine and its environment.

Beyond critical systems, many areas could benefit from \textit{forbidden actions}. In Natural Language Generation \cite{reiter1997building} or Dialogue systems \cite{chandramohan2010optimizing, chen2017survey}, syntax parser or auto-correct mechanism can act as an external rejection signal. Indicating which word does not fit a generated sentence or pointing out grammar mistakes could improve language generation by greatly reducing word's space, leveraging language learning and generation.

These examples show a potential misalignment between the standard RL frameworks and the potential real-world applications. Of course, designing constraints to avoid critical situations requires expert knowledge about the system to be controlled. But it is often the case that environments already implement such contingency measures (obstacle avoidance, circuit breaker, etc.). 

In this paper, we consider a simple type of external constraints, prevalent in many real-world problems: when the agent is about to perform a hazardous action, the system rejects it and thus prevents the agent from doing so. The agent then follows the natural dynamics of the environment \citep{alshiekh2018safe}. We aim at building an algorithm that learns from these rejected actions. 

In the general Markov Decision Process (MDP) framework \citep{puterman2014markov}, a rejected action, from the agent's point of view, would be seen as a transition to the same state. Everything happens like if the action had no effect. This misrepresents the potential harmfulness of the action and prevents the agent from learning anything useful from the rejected action (See \autoref{subsec:mdp-f} for a detailed explanation). We want a model-free reinforcement learning to benefit from those constraints so as to learn faster to avoid hazardous states and alleviate exploration problems.


In the coming sections, after introducing the RL paradigm, we propose a constrained version of Deep $Q$-learning (DQN)~\citep{mnih2015human} by adding a classification loss that maintains $Q$-values of forbidden actions below valid ones. We then validate our method empirically, showing that vanilla DQN struggles at solving tasks with rejected actions while our algorithm reduces the number of calls to forbidden actions. It accelerates convergence to near-optimal policies compared to standard DQN. Experiments are conducted on two tasks: A maze navigation using visual features and a text-based game. 

\section{Context: Reinforcement Learning}\label{subsec:rl}


In the reinforcement learning (RL) paradigm~\citep{sutton2018reinforcement}, an agent learns to interact with its environment so as to maximize a cumulative function of rewards. The environment to be controlled is modeled as a Markov Decision Process (MDP) that is a tuple $\{S,A,\mathcal{P},\mathcal{R},\gamma\}$ which elements are defined as follows. 
At each time step $t$, the agent is in a state $s_{t} \in \mathcal{S}$, where it selects an action $a_t \in \mathcal{A}$ according to its policy $\pi: \mathcal{S} \rightarrow \mathcal{A}$. 
It moves to state $s_{t+1}$ according to a transition kernel $\mathcal{P}$ and receives a reward $r_t=r(s_t,a_t)$ drawn from the environment's reward function $\mathcal{R}: \mathcal{S} \times \mathcal{A} \rightarrow \mathds{R}$.
The quality of the policy is assessed by the $Q$-function defined by  
$$Q^{\pi}(s,a) = \mathds{E}_{\pi} \left[ \sum_{t} \gamma^{t} r(s_{t},a_{t}) |  s_{0}=s, a_{0}=a\right]$$ 

\noindent for all $(s,a)$ where $\gamma \in [0,1]$ is the discount factor. 
The optimal $Q$-value is defined as $Q^{*}(s,a) = \max_{\pi}Q^{\pi}(s,a)$, from which the optimal policy $\pi^{*}$ is derived. 

$$\pi^*(s) \in \text{arg}\max_a Q^*(s,a)$$


We here use the Deep $Q$-learning (DQN) algorithm~\citep{mnih2015human} to approximate the optimal $Q$-function with neural networks and perform off-policy updates by sampling transitions $(s_{t}, a_{t}, r_{t}, s_{t+1})$ from a replay buffer~\citep{lin1992self}. 




\section{Method}\label{subsec:methods}

\subsection{Feedback Signal and MDP-$\mathcal{F}$}\label{subsec:mdp-f}

In this section, we present a way to integrate forbidden actions into the MDP framework.
We augment the MDP model with a \textit{\fs}~\citep{alshiekh2018safe}, a Boolean indicating whether an action was accepted by the environment or rejected. A MDP-$\mathcal{F}$ is then defined as a tuple $<\mathcal{S}, \mathcal{A}, \mathcal{P}, \mathcal{R}, \gamma, \mathcal{F}>$ where $\mathcal{F}$ is a function mapping a state $s_t$ and action $a_t$ to a binary value. $$\mathcal{F}: S \times \mathcal{A} \rightarrow (0,1)$$ with 0 meaning the action is valid and 1 meaning unsafe/rejected action.

Vanilla $Q$-learning struggles to differentiate between actions flagged as forbidden and valid ones.
Consider the following example: an agent in a state $s$ takes action $a$ flagged as forbidden ($\mathcal{F}(a,s) = f = 1$). When applying the $Q$-learning update ($Q(s,a_{feed}) = r(s,a_{feed}) + \gamma max Q(s',a')$), since the action was rejected, $r(s,a) = 0$ and $s=s'$. Thus the update becomes $Q(s,a_{feed}) = \gamma max Q(s,a')$. In current Deep Reinforcement Learning setup $\gamma$ is usually set between $0.99$ \citep{mnih2015human} and $0.999$ \citep{pohlen2018observe}. DQN-like algorithm will require lots of transitions to make the $Q$-function of forbidden actions smaller, thus loosing time to explore and collect useful samples.
We emphasize that an invalid action indicates an action that could be harmful, so rapidly identifying and avoiding those potentially dangerous situations is crucial.

\subsection{Frontier loss}
We take inspiration from the learning from demonstrations paradigm where one wants to use expert demonstrations to induce the usage of preferred actions in RL agents. 

\paragraph{Expert loss and Imitation learning} In Imitation learning, few expert demonstrations are available and extracting as much information from those is essential. For example \citep{hester2018deep, Piot2014} slightly modify the $Q$-learning update to nudge expert actions-value above other actions. This is done by adding a secondary loss inspired by structured classification:
$$\text{Minimize\ }J_{\text{E}}(Q) = \max_{a \in A}[Q(s, a) +l(a^E, a)] - Q(s, a^E) $$ 
where $a^E$ is the action of the expert, $l(a^E,a) = 0$ when $a^E = a$ and $m$ otherwise. This nudges the $Q$-value of actions taken by the expert above the $Q$-value of other actions by at most a certain margin $m$.

Similarly, we want to derive a loss that penalizes the $Q$-function when a \textit{forbidden action's} value excesses the value of a valid one. 

\paragraph{Frontier loss}

The optimal policy $\pi^*$ (derived from $Q^*$) will never take a forbidden action as it keeps the agent in the same state. Based on this assumption we can derive the following rule: 
for every state encountered during training, the $Q$-values of all forbidden actions should be below the one of each valid actions, within a certain margin $m$.
This defines a new loss we want to minimize that we call \textit{frontier loss} $J_\mathcal{F}$ : 
\begin{figure}[H]
  \centering
  \begin{equation}
     J_{\mathcal{F}}(Q) = Q(s, a^-) - \min_{a \in \mathcal{V}_\text{s} }[Q(s, a) - m] 
  \end{equation}
  \begin{tabular}{ll}
    $\text{where } \mathcal{V}_\text{s}$ = & $\mathcal{A}(s) \cap \{a \text{ s.t. } \mathcal{F}(s,a)=0\}$\\
    and\hspace{0.2em} $a^- \in $ & $\mathcal{A}(s) \cap \{a \text{ s.t. } \mathcal{F}(s,a)=1\}$
  \end{tabular}
\end{figure}
\begin{figure}
\centering
\includegraphics[width=0.75\columnwidth]{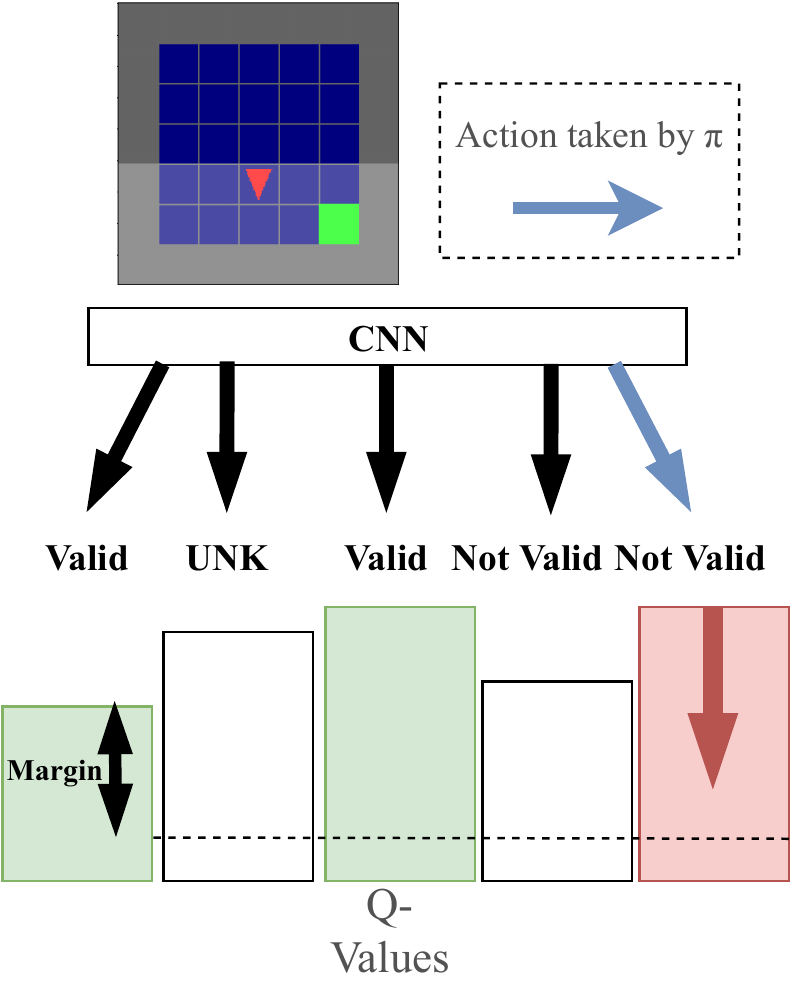}
\caption{Illustration of frontier loss.}
\label{fig:illuLoss}
\end{figure}
The margin $m$ is an hyper-parameter that depends mostly on the $Q$-values magnitude. In our experiments, since the rewards are bounded between 0 and 1, the margin is small ($m=0.1$).

\begin{figure}
\centering
\includegraphics[width=0.98\columnwidth]{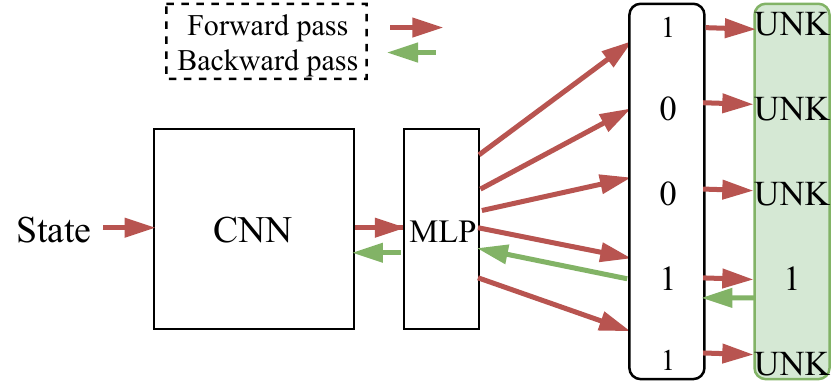}
\caption{Training procedure: the model predicts the validity for each action, and we only backpropagate for the action the agent took.}
\label{fig:action_classif_model}
\end{figure}

\paragraph{Frontier loss and classification} The main problem regarding this objective function is the need to know which actions are valid for every state. In most tasks, it's unlikely that the agent visits a specific state more than once (\textit{e.g.} visual domains). Thus, function approximation is required to estimate which actions are valid in a given state. 
To achieve this, we train a neural network to predict, for each state, which action will be \textbf{valid}. Along agent's trajectories, for every action taken, we store the corresponding feedback, creating a dataset of $(s, a, f)$. The network, taking the state as input, predicts a binary value for every action (0 for valid, 1 for invalid). For each state in our transition dataset, since most of the time only one action is labelled, we need to adapt the training regime. We can achieve this by masking the gradients from untaken action, only backpropagating for the action the policy $\pi$ took. The training procedure is illustrated on \autoref{fig:action_classif_model}. 

To consider an action as valid and to avoid early mis-classifications, we put a threshold after the \textit{sigmoid} function. The action is considered to be valid if its activation is above the threshold. 

\paragraph{DQN-$\mathcal{F}$} We construct a new algorithm, DQN-$\mathcal{F}$, that simply combines the frontier loss and Deep $Q$-learning. We build a composite loss by using weighting factors $\lambda_\text{DQN}$ and $\lambda_\mathcal{F}$ to balance the DQN and the frontier losses. For all the experiments described below, we use $\lambda_{\text{DQN}}=1$ and $\lambda_\mathcal{F} = 0.5$. Not much tweaking is required regarding these hyper-parameter.

$$ J(Q) = \lambda_{\text{DQN}} J_{DQN}(Q) + \lambda_\mathcal{F} J_{\mathcal{F}}(Q)$$

\begin{algorithm}
 \KwData{minibatch $b$ from replay buffer $\mathcal{R}$, Q-network $\mathcal{Q}$, classification network $\mathcal{C}$}
 \KwResult{Frontier loss}
 loss = 0\;
 \For{$($state $s$, action $a$, feedback $f)$ in minibatch $b$}{
  \If{f = 1}{
   $a^- \gets$ a \algorithmiccomment{Renaming for clarity}\\
   actions$_{valid} \gets$ C(s); \algorithmiccomment{Estimated valid action set}\\
   \If{$\min$[Q(s, actions$_{valid}$)] < Q(s, $a^-$) - m}{
        loss = loss + $|| \min$[Q(s, actions$_{valid}$)]\\ - Q(s, $a^-$) -m $||^2$}
   }
}
 return loss;
 \caption{Frontier loss and classification network.}
\end{algorithm}

\section{Experiments}

We assess our method on two different tasks: a toy problem (MiniGrid) and a text-based game (TextWorld).

\subsection{MiniGrid Enviroment}

The first environment is a simple visual gridworld presented in \citep{chevalier2019baby}. 
The goal is to reach the green zone starting from a random point. Since we want to study how the agent can integrate feedback about action's validity, we increase the action space size. To do so, we create $k$ different room types where the color of the background indicates \textit{which set of actions is valid}. The primary action space is composed of 3 actions (Turning Left, Turning Right, Going Forward) for navigation, but each action is duplicated $k$ times. The action space size becomes $3 \times k$ but only 3 are valid in a given room. For example, in the red room, only actions 11, 12, 13 are valid, and all the others are returning a \textbf{not valid} feedback. In our setup, we use $k=5$ making a total of 15 actions.

\begin{figure}
\centering
\includegraphics[width=0.35\columnwidth]{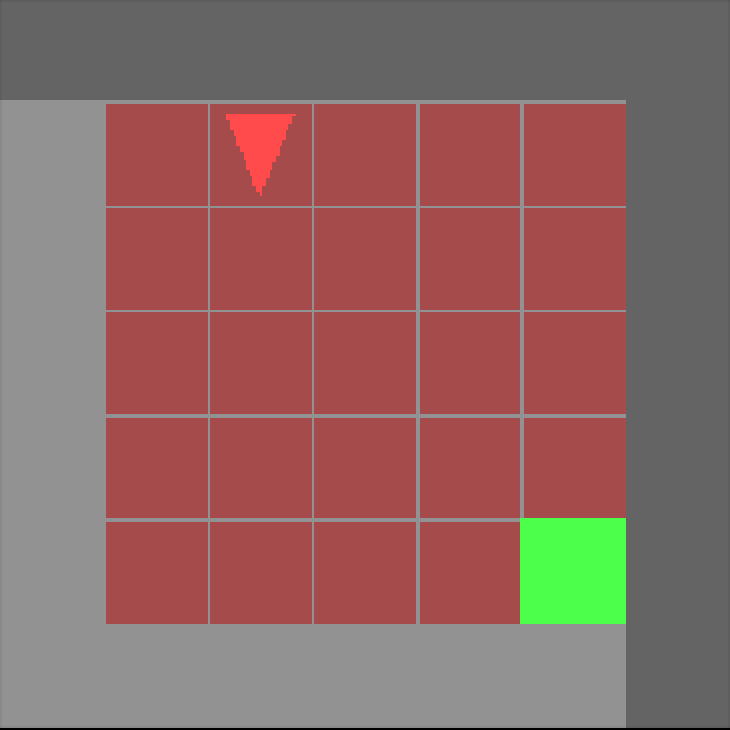}
\caption{An instance of the MiniGrid problem. The state is a partial view of the maze (point of view of the agent) to avoid problem regarding partial observability, we stacked the last 3 frames.}
\label{fig:minigrid_ex}
\end{figure}

The state space is an embedding of the agent's point of view represented as different features maps allowing the use of convolution layers (more details in \citep{chevalier2019baby}). Since the environment is partially observable, we stack the last three frames as in \citet{mnih2015human} but we do not use frame-skipping. An episode ends when the agent reaches the green zone or after 200 environment steps as illustrated on \autoref{fig:minigrid_ex}.

\subsection{TextWorld Environment}

\begin{figure}
\centering
\includegraphics[width=\columnwidth]{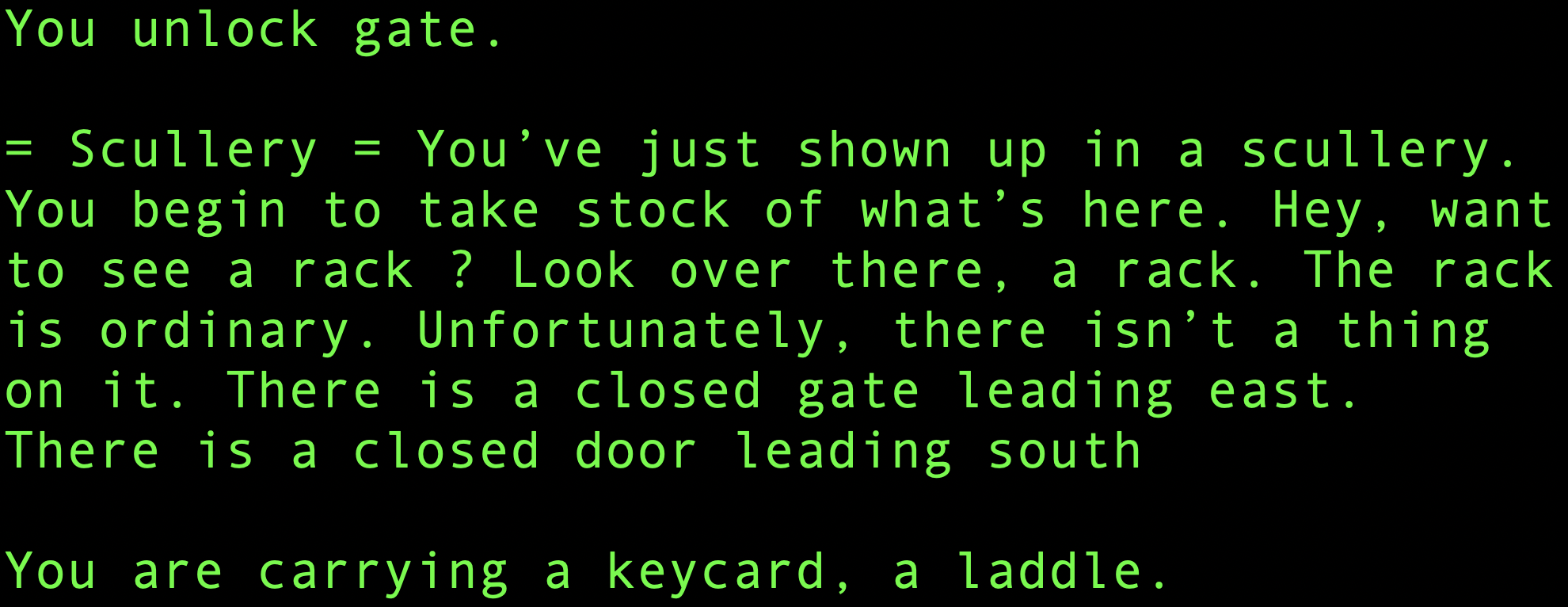}
\caption{An example of interaction in TextWorld. The agent has access to: what happened after its latest action ("You open the door, it’s very dark in here, [...]"), a room description ("Attic, an empty room, maybe you should head back. You can go North, East") and its inventory content ("Keycard, Mask").}
\label{fig:text_word_example}
\end{figure}

TextWorld \citep{cote18textworld} is a text-based game where the agent interacts with the environment using short sentences. We generated a game composed of 3 rooms, 7 objects, and quest length of size 4. An example is shown on \autoref{fig:text_word_example}. In this context, we modified the environment to fit our needs. The action space is composed of all <action> <object> pairs, creating a total of 46 actions. Most of the actions created will be rejected by the simulator since they will not fit the situation the agent is facing. For example, the action "take sword" will be rejected if no sword is available. An illustration of the game can be found \autoref{fig:text_word_example}.

\subsection{Model and architecture}

During all experiments, we use Double Deep $Q$-Network (DQN) \citep{hasselt2016deep} with uniform Experience Replay and $\epsilon$-greedy exploration. In the Minigrid environment, we use a Convolution Neural Network \citep{lecun1995convolutional} with a fully-connected layer on top. In TextWorld, inventory, observation, and room descriptions are each encoded by an LSTM \citep{hochreiter1997long} processed by a fully-connected layer on top. 

The classification network matches exactly the architecture used by DQN, \textit{i.e.} ConvNet for Minigrid and LSTM's for TextWorld, the only difference resides in training (explained \autoref{fig:action_classif_model})

\section{Results}

\begin{figure}
  \begin{minipage}{\columnwidth}
    \centering
    \includegraphics[width=\columnwidth]{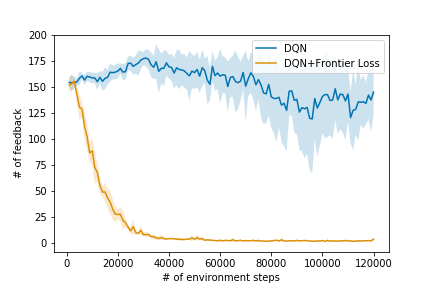}
    \caption{\textbf{Minigrid results: Number of times a forbidden action is taken.} DQN-$\mathcal{F}$ (yellow) DQN (blue). Results are averaged over five random seeds. The shaded area represents one standard deviation.}
    \label{fig:minigrid_feed}
  \end{minipage}

  \begin{minipage}{\columnwidth}
    \centering
    \includegraphics[width=\columnwidth]{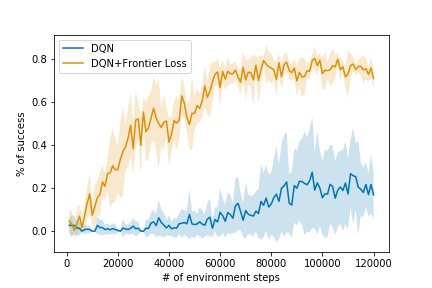}
    \caption{\textbf{Minigrid results: Percentage of success over time.} DQN-$\mathcal{F}$ (yellow) DQN (blue). Results are averaged over 5 random seeds. The shaded area represents one standard deviation.}
    \label{fig:minigrid_reward}
  \end{minipage}
\end{figure}

\begin{figure}
  \begin{minipage}{\columnwidth}
    \centering
    \includegraphics[width=\columnwidth]{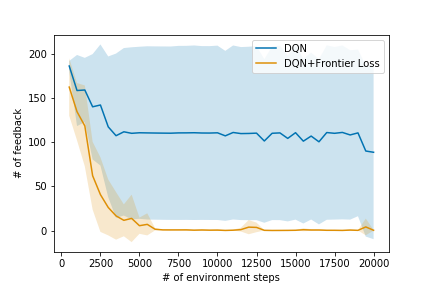}
    \caption{\textbf{Textworld results: Number of invalid action taken by the agent.} DQN-$\mathcal{F}$ (yellow) DQN (blue). Results are averaged over nine random seeds. The shaded area represents one standard deviation.}
    \label{fig:text_feedback}
  \end{minipage}

  \begin{minipage}{\columnwidth}
    \centering
    \includegraphics[width=\columnwidth]{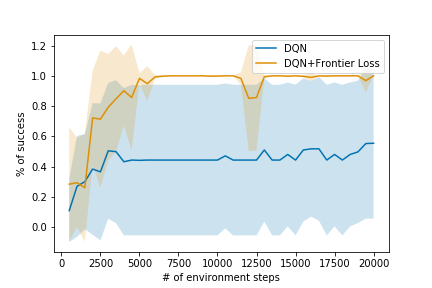}
    \caption{\textbf{Textworld results: Percentage of success over time.} DQN-$\mathcal{F}$ (yellow) DQN (blue). Results are averaged over 9 random seeds. The shaded area represents one standard deviation.}
    \label{fig:text_reward}
  \end{minipage}
\end{figure}

\begin{figure}
    \begin{minipage}{0.49\columnwidth}
    \centering
    \includegraphics[width=\columnwidth]{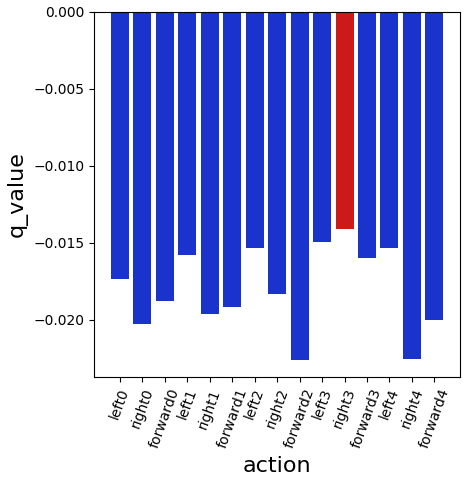}
    \end{minipage}
    \begin{minipage}{0.49\columnwidth}
    \centering
    \includegraphics[width=\columnwidth]{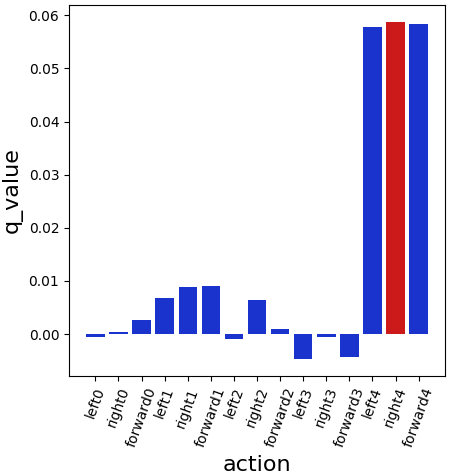}
    
    \end{minipage}
\caption{\textbf{Q-values, after 150 episodes of training (\numprint{30000}) env steps) for certain state}. Left: DQN; Right: DQN with frontier loss. We can clearly see how the frontier-loss shapes Q-values, clearly separating forbidden action from good ones. This cherry-picked example illustrates well how the $Q$-distribution is modified, even at that early training stage.}
\label{fig:earlyq}
\end{figure}

\begin{figure}
    \begin{minipage}{0.49\columnwidth}
    \centering
    \includegraphics[width=\columnwidth]{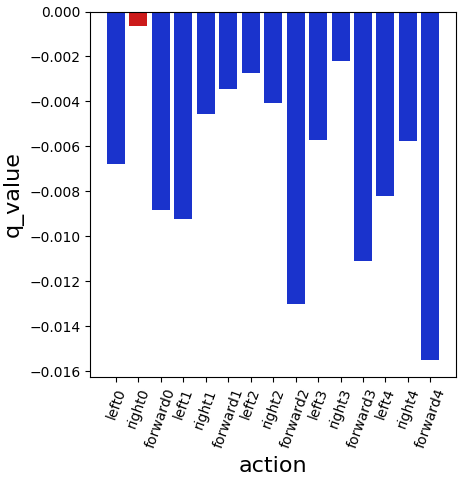}
    \end{minipage}
    \begin{minipage}{0.49\columnwidth}
    \centering
    \includegraphics[width=\columnwidth]{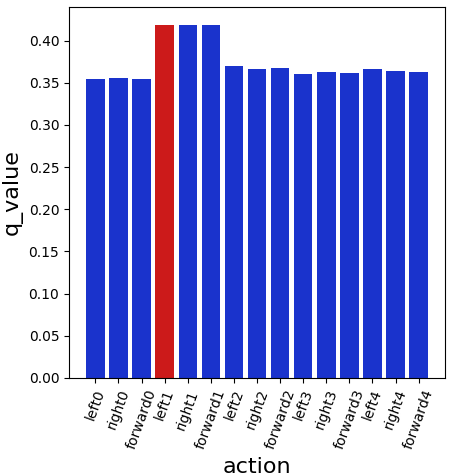}
    \end{minipage}
\caption{\textbf{Q-values, after \numprint{100000} training steps} Left: DQN; Right: DQN-$\mathcal{F}$. Even after \numprint{100000} network updates, DQN still struggles to differentiate good action from bad ones. This cherry-picked example illustrates well how the $Q$-distribution is modified and how $Q$-values are kept separated even later during training.}
\label{fig:endq}
\end{figure}

In \autoref{fig:minigrid_feed}, \autoref{fig:minigrid_reward}, \autoref{fig:text_feedback}, \autoref{fig:text_reward},  we compare DQN and DQN-$\mathcal{F}$. In the Minigrid domain, DQN struggles to find the optimal policy and reaches only 20\% of the time the exit. Most of the time, DQN is able to solve one room but fails to find the set of actions for each room, performing forbidden actions over and over. On the contrary, the frontier loss is guiding DQN-$\mathcal{F}$, reducing the number of negative feedback signals from the environment, and helping to find the optimal policy. Those results are echoed in the TextWorld experiment. DQN solves the game half of the time, and the other half does not encounter the reward and as a result, can not solve the game. This could be mitigated by having a better exploration strategy, but it shows that shaping $Q$-values with the frontier loss is enough to reduce the sample complexity and guide exploration. 
We want to emphasize that in early stages of the training, the classification network performs poorly due to low quantity of samples but it does not hurt the performances of DQN-$\mathcal{F}$, it's able to quickly learn to avoid forbidden actions. Visualization of $Q$-values at different stages of training can be found on \autoref{fig:earlyq}, \autoref{fig:endq}. They clearly illustrate the benefits of using the frontier loss in those setups. Even in the early training stage, the separation between valid actions and invalid is clear, alleviating the difficulty of finding the optimal policy. Where as DQN $Q$-values are really hard to distinguish from each other.

\section{Related Works}

\paragraph{Action Elimination}
Closely related to our work is the notion of \textit{action elimination} which was introduced in \citep{even2006action}. The main idea developed in that work, applied to Multi-Arm Bandits \citep{lattimore2018bandit, lai1985asymptotically, robbins1952some} is to get rid of a sub-optimal action as soon as the value of this action is out of some confidence interval. 

A similar idea was applied in Deep Reinforcement Learning by \cite{zahavy2018learn}. This article shares similarities with ours as the authors are trying to eliminate actions based on a signal given by the environment, indicating if the action is valid or not. They are using a contextual bandit to assess the elimination signal's certainty, and remove actions from the action set $A$ when the confidence is above a certain threshold. 
The main difference is in the way the elimination signal acts on $Q$-learning. In their case, the elimination signal doesn't change $Q$-values but modifies the action set directly. They need extra care because removing a valid action can irreversibly damage the policy where as in our case, we only nudge values.

\cite{alshiekh2018safe} define the term \textit{shielding} similar to our notion of feedback. The simulator rejects potentially harmful actions. To learn from this process, the agent outputs a set of actions, ordered by preferences, and the simulator picks the best-allowed action.

\paragraph{Learning when the environment takes over} \citet{orseau2016safely} design agent to \textit{not} take into account feedback from the environment. For example, for an agent operating in real-time, it may be necessary for a human operator to prevent executing a harmful sequence and lead the agent into a safer situation.  However, if the learning agent expects to receive rewards from this sequence, it may learn in the long run to avoid such interruptions, for example, by disabling the \textit{off-button}. Under this setup, they showed that $Q$-learning could be interrupted safely, supporting our hypothesis that $Q$-learning is not integrating the feedback signal.

\paragraph{Reward Shaping}\citet{lipton2016combating} introduced the notion of Intrinsic Fear (IF) to deal with dangerous states. Deep agents tend to periodically revisit these states upon forgetting their existence under a new policy. IF is shaping the reward to guard agents against periodic catastrophes. To achieve that they train a model to predict the probability of imminent dangerous state and it penalizes the true reward function $R$ with an intrinsic reward $F$ 

\paragraph{Large Discrete Action Space} A benefit of our method is to simplify policy space search when using bigger action space. Dealing with large discrete action space in Deep Reinforcement Learning was studied by \cite{dulac2015deep} where they use an action embedding in continuous space and map it to a discrete action. Recent methods build upon this by also learning the action embedding instead of using a pre-computed one \cite{chandak2019learning, tennenholtz2019natural, chen2019learning, adolphs2019ledeepchef}
Another body of literature explores how to reduce combinatorial action space by using an encoding dictionary \citep{dulac2012fast} rendering problems with large action sets tractable. \citep{he2016combi, he2016natural} are dealing with variable action space, meaning that at every time step, the action set size is different. To cope with this challenge, they compute a state embedding and action embedding for every action available then perform a dot product between the two. This allows them to deal with variable sized action set. 

\section{Conclusion}

Critical real-world systems are constrained by safety measures that prevent hazardous action. We hypothesize that Q-learning, a model-free reinforcement learning method struggles with those constraints. 
In this paper, we proposed a frontier loss, combined with a classification network, to help DQN. This algorithm nudges rejected actions $Q$-values below $Q$-values of valid actions. We showed empirically that the frontier loss reduces the number of calls to rejected actions and guides early exploration, helping Deep $Q$-learning achieving higher performances. We demonstrate its effectiveness on two benchmarks, a visual grid world and a TextWorld domain. 

\paragraph{Limitations and Future Work}

At the moment, applying the frontier to continuous action space is non-trivial but it would be a key to use this type of algorithm in robotics and more realistic settings. Another future improvement would be to combine the loss with action's embedding could allow generalization to unseen actions. For example, learning that "Take sword" is rejected "Grab sword" shouldn't be considered by the algorithm.

\subsubsection*{Acknowledgments}

The authors would like to acknowledge the stimulating research environment of the SequeL INRIA Project-Team. Special thanks to Florian Strub and Edouard Leurent for fruitful discussions. 
We acknowledge the following agencies for research funding and computing support: Project BabyRobot (H2020-ICT-24-2015, grant agreement no.687831), and CPER Nord-Pas de Calais/FEDER DATA Advanced data science and technologies 2015-2020

\bibliography{ref}

\begin{thebibliography}{}

\bibitem[Adolphs and Hofmann, 2019]{adolphs2019ledeepchef}
Adolphs, L. and Hofmann, T. (2019).
\newblock Ledeepchef: Deep reinforcement learning agent for families of
  text-based games.
\newblock {\em In NeurIPS Deep Reinforcement Learning Workshop}.

\bibitem[Alshiekh et~al., 2018]{alshiekh2018safe}
Alshiekh, M., Bloem, R., Ehlers, R., K{\"o}nighofer, B., Niekum, S., and Topcu,
  U. (2018).
\newblock Safe reinforcement learning via shielding.
\newblock In {\em Proc. of AAAI}.

\bibitem[Bellemare et~al., 2017]{bellemare2017distributional}
Bellemare, M.~G., Dabney, W., and Munos, R. (2017).
\newblock A distributional perspective on reinforcement learning.
\newblock In {\em Proceedings of the 34th International Conference on Machine
  Learning-Volume 70}, pages 449--458. JMLR. org.

\bibitem[Chandak et~al., 2019]{chandak2019learning}
Chandak, Y., Theocharous, G., Kostas, J., Jordan, S., and Thomas, P. (2019).
\newblock Learning action representations for reinforcement learning.
\newblock In {\em Proc. of ICML}.

\bibitem[Chandramohan et~al., 2010]{chandramohan2010optimizing}
Chandramohan, S., Geist, M., and Pietquin, O. (2010).
\newblock Optimizing spoken dialogue management with fitted value iteration.
\newblock In {\em Proc. of ISCA}.

\bibitem[Chen et~al., 2017]{chen2017survey}
Chen, H., Liu, X., Yin, D., and Tang, J. (2017).
\newblock A survey on dialogue systems: Recent advances and new frontiers.
\newblock {\em Acm Sigkdd Explorations Newsletter}, 19(2):25--35.

\bibitem[Chen et~al., 2019]{chen2019learning}
Chen, Y., Chen, Y., Yang, Y., Li, Y., Yin, J., and Fan, C. (2019).
\newblock Learning action-transferable policy with action embedding.
\newblock In {\em Proc. of AAAI}.

\bibitem[Chevalier-Boisvert et~al., 2019]{chevalier2019baby}
Chevalier-Boisvert, M., Bahdanau, D., Lahlou, S., Willems, L., Saharia, C.,
  Nguyen, T.~H., and Bengio, Y. (2019).
\newblock Baby{AI}: First steps towards grounded language learning with a human
  in the loop.
\newblock In {\em Proc. of ICLR}.

\bibitem[C\^ot\'e et~al., 2018]{cote18textworld}
C\^ot\'e, M.-A., K\'ad\'ar, A., Yuan, X., Kybartas, B., Barnes, T., Fine, E.,
  Moore, J., Hausknecht, M., Asri, L.~E., Adada, M., Tay, W., and Trischler, A.
  (2018).
\newblock Textworld: A learning environment for text-based games.
\newblock {\em CoRR}, abs/1806.11532.

\bibitem[Daubigney et~al., 2011]{daubigney2011uncertainty}
Daubigney, L., Ga{\v{s}}i{\'c}, M., Chandramohan, S., Geist, M., Pietquin, O.,
  and Young, S. (2011).
\newblock Uncertainty management for on-line optimisation of a pomdp-based
  large-scale spoken dialogue system.

\bibitem[Dulac-Arnold et~al., 2012]{dulac2012fast}
Dulac-Arnold, G., Denoyer, L., Preux, P., and Gallinari, P. (2012).
\newblock Fast reinforcement learning with large action sets using
  error-correcting output codes for mdp factorization.
\newblock In {\em Proc. of ECML and PKDD}. Springer.

\bibitem[Dulac-Arnold et~al., 2015]{dulac2015deep}
Dulac-Arnold, G., Evans, R., van Hasselt, H., Sunehag, P., Lillicrap, T., Hunt,
  J., Mann, T., Weber, T., Degris, T., and Coppin, B. (2015).
\newblock Deep reinforcement learning in large discrete action spaces.
\newblock {\em arXiv preprint arXiv:1512.07679}.

\bibitem[Dulac-Arnold et~al., 2019]{dulac2019challenges}
Dulac-Arnold, G., Mankowitz, D., and Hester, T. (2019).
\newblock Challenges of real-world reinforcement learning.
\newblock In {\em Real-world Sequential Decision Making Workshop @ ICML 2019}.

\bibitem[El-Tantawy et~al., 2013]{el2013multiagent}
El-Tantawy, S., Abdulhai, B., and Abdelgawad, H. (2013).
\newblock Multiagent reinforcement learning for integrated network of adaptive
  traffic signal controllers (marlin-atsc): methodology and large-scale
  application on downtown toronto.
\newblock {\em Proc. of TITS}.

\bibitem[Even-Dar et~al., 2006]{even2006action}
Even-Dar, E., Mannor, S., and Mansour, Y. (2006).
\newblock Action elimination and stopping conditions for the multi-armed bandit
  and reinforcement learning problems.
\newblock {\em Journal of machine learning research}, 7(Jun):1079--1105.

\bibitem[Geist and Pietquin, 2011]{geist2011managing}
Geist, M. and Pietquin, O. (2011).
\newblock {Managing uncertainty within the KTD framework}.
\newblock In {\em Active Learning and Experimental Design workshop @ AISTATS
  2010}, pages 157--168.

\bibitem[Gu et~al., 2017]{gu2017deep}
Gu, S., Holly, E., Lillicrap, T., and Levine, S. (2017).
\newblock Deep reinforcement learning for robotic manipulation with
  asynchronous off-policy updates.
\newblock In {\em Proc. of ICRA}. IEEE.

\bibitem[Hashimoto et~al., 2018]{hashimoto2018artificial}
Hashimoto, D.~A., Rosman, G., Rus, D., and Meireles, O.~R. (2018).
\newblock Artificial intelligence in surgery: promises and perils.
\newblock {\em Annals of surgery}, 268(1):70.

\bibitem[Hasselt et~al., 2016]{hasselt2016deep}
Hasselt, H.~v., Guez, A., and Silver, D. (2016).
\newblock Deep reinforcement learning with double q-learning.
\newblock In {\em Proc. of AAAI}.

\bibitem[He et~al., 2016a]{he2016natural}
He, J., Chen, J., He, X., Gao, J., Li, L., Deng, L., and Ostendorf, M. (2016a).
\newblock Deep reinforcement learning with a natural language action space.
\newblock In {\em Proc. of ACL}.

\bibitem[He et~al., 2016b]{he2016combi}
He, J., Ostendorf, M., He, X., Chen, J., Gao, J., Li, L., and Deng, L. (2016b).
\newblock Deep reinforcement learning with a combinatorial action space for
  predicting popular reddit threads.
\newblock In {\em Proc. of EMNLP}.

\bibitem[Hester et~al., 2018]{hester2018deep}
Hester, T., Vecerik, M., Pietquin, O., Lanctot, M., Schaul, T., Piot, B.,
  Horgan, D., Quan, J., Sendonaris, A., Osband, I., et~al. (2018).
\newblock {Deep Q-learning from demonstrations}.
\newblock In {\em Proc. of AAAI}.

\bibitem[Hochreiter and Schmidhuber, 1997]{hochreiter1997long}
Hochreiter, S. and Schmidhuber, J. (1997).
\newblock Long short-term memory.
\newblock {\em Neural computation}, 9(8):1735--1780.

\bibitem[Lai and Robbins, 1985]{lai1985asymptotically}
Lai, T.~L. and Robbins, H. (1985).
\newblock Asymptotically efficient adaptive allocation rules.
\newblock {\em Advances in applied mathematics}, 6(1):4--22.

\bibitem[Lattimore and Szepesv{\'a}ri, 2018]{lattimore2018bandit}
Lattimore, T. and Szepesv{\'a}ri, C. (2018).
\newblock Bandit algorithms.

\bibitem[LeCun et~al., 1995]{lecun1995convolutional}
LeCun, Y., Bengio, Y., et~al. (1995).
\newblock Convolutional networks for images, speech, and time series.
\newblock {\em The handbook of brain theory and neural networks}.

\bibitem[Lemon and Pietquin, 2012]{lemon2012data}
Lemon, O. and Pietquin, O. (2012).
\newblock {\em Data-driven methods for adaptive spoken dialogue systems:
  Computational learning for conversational interfaces}.
\newblock Springer Science \& Business Media.

\bibitem[Leurent and Mercat, 2019]{leurent2019social}
Leurent, E. and Mercat, J. (2019).
\newblock Social attention for autonomous decision-making in dense traffic.
\newblock In {\em Machine Learning for Autonomous Driving Workshop at NeurIPS
  2019}.

\bibitem[Levine et~al., 2016]{levine2016end}
Levine, S., Finn, C., Darrell, T., and Abbeel, P. (2016).
\newblock End-to-end training of deep visuomotor policies.
\newblock {\em The Journal of Machine Learning Research}, 17(1):1334--1373.

\bibitem[Lin, 1992]{lin1992self}
Lin, L.-J. (1992).
\newblock Self-improving reactive agents based on reinforcement learning,
  planning and teaching.
\newblock {\em Machine learning}, 8(3-4):293--321.

\bibitem[Lipton et~al., 2016]{lipton2016combating}
Lipton, Z.~C., Azizzadenesheli, K., Kumar, A., Li, L., Gao, J., and Deng, L.
  (2016).
\newblock Combating reinforcement learning's sisyphean curse with intrinsic
  fear.
\newblock {\em arXiv preprint arXiv:1611.01211}.

\bibitem[Mao et~al., 2016]{mao2016resource}
Mao, H., Alizadeh, M., Menache, I., and Kandula, S. (2016).
\newblock Resource management with deep reinforcement learning.
\newblock In {\em Proc. of ACM Workshop on Hot Topics in Networks}.

\bibitem[Mnih et~al., 2015]{mnih2015human}
Mnih, V., Kavukcuoglu, K., Silver, D., Rusu, A.~A., Veness, J., Bellemare,
  M.~G., Graves, A., Riedmiller, M., Fidjeland, A.~K., Ostrovski, G., et~al.
  (2015).
\newblock Human-level control through deep reinforcement learning.
\newblock {\em Nature}, 518(7540):529.

\bibitem[Orseau and Armstrong, 2016]{orseau2016safely}
Orseau, L. and Armstrong, S. (2016).
\newblock Safely interruptible agents.
\newblock In {\em Proc. of UAI}.

\bibitem[O’Donoghue et~al., 2018]{o2018uncertainty}
O’Donoghue, B., Osband, I., Munos, R., and Mnih, V. (2018).
\newblock The uncertainty bellman equation and exploration.
\newblock In {\em International Conference on Machine Learning}, pages
  3836--3845.

\bibitem[Piot et~al., 2014]{Piot2014}
Piot, B., Geist, M., and Pietquin, O. (2014).
\newblock {Boosted Bellman residual minimization handling expert
  demonstrations}.
\newblock {\em Lecture Notes in Computer Science (including subseries Lecture
  Notes in Artificial Intelligence and Lecture Notes in Bioinformatics)}, 8725
  LNAI(PART 2):549--564.

\bibitem[Pohlen et~al., 2018]{pohlen2018observe}
Pohlen, T., Piot, B., Hester, T., Azar, M.~G., Horgan, D., Budden, D.,
  Barth-Maron, G., Van~Hasselt, H., Quan, J., Ve{\v{c}}er{\'\i}k, M., et~al.
  (2018).
\newblock Observe and look further: Achieving consistent performance on atari.
\newblock {\em arXiv preprint arXiv:1805.11593}.

\bibitem[Puterman, 2014]{puterman2014markov}
Puterman, M.~L. (2014).
\newblock {\em Markov Decision Processes.: Discrete Stochastic Dynamic
  Programming}.
\newblock John Wiley \& Sons.

\bibitem[Reiter and Dale, 1997]{reiter1997building}
Reiter, E. and Dale, R. (1997).
\newblock Building applied natural language generation systems.
\newblock {\em Natural Language Engineering}, 3(1):57--87.

\bibitem[Robbins, 1952]{robbins1952some}
Robbins, H. (1952).
\newblock Some aspects of the sequential design of experiments.
\newblock {\em Bulletin of the American Mathematical Society}, 58(5):527--535.

\bibitem[Sallab et~al., 2017]{sallab2017deep}
Sallab, A.~E., Abdou, M., Perot, E., and Yogamani, S. (2017).
\newblock Deep reinforcement learning framework for autonomous driving.
\newblock {\em Electronic Imaging}.

\bibitem[Sutton and Barto, 2018]{sutton2018reinforcement}
Sutton, R.~S. and Barto, A.~G. (2018).
\newblock {\em Reinforcement learning: An introduction}.

\bibitem[Tennenholtz and Mannor, 2019]{tennenholtz2019natural}
Tennenholtz, G. and Mannor, S. (2019).
\newblock The natural language of actions.
\newblock In {\em Proc. of ICML}.

\bibitem[Wang and Xu, 2012]{wang2012dam}
Wang, R. and Xu, L. (2012).
\newblock Multi-agent dam management model based on improved reinforcement
  learning technology.
\newblock {\em Applied Mechanics and Materials}, 198.

\bibitem[Zahavy et~al., 2018]{zahavy2018learn}
Zahavy, T., Haroush, M., Merlis, N., Mankowitz, D.~J., and Mannor, S. (2018).
\newblock Learn what not to learn: Action elimination with deep reinforcement
  learning.
\newblock In {\em Proc. of NeurIPS}.

\bibitem[Zhou et~al., 2017]{zhou2017optimizing}
Zhou, Z., Li, X., and Zare, R.~N. (2017).
\newblock Optimizing chemical reactions with deep reinforcement learning.
\newblock {\em ACS central science}, 3(12):1337--1344.

\end{thebibliography}
\bibliographystyle{apalike}
\small

\appendix
\section{Appendix A: Training details}\label{appendix:training}

\begin{table}[h]
\begin{tabular}{|l|l|}
\hline
\textbf{Convolution Layer}               & 16, 32, 64           \\ \hline
\textbf{Kernel size}                     & 2,2,2                \\ \hline
\textbf{Pooling}                         & 2 on the first layer \\ \hline
\textbf{Fully Connected hidden size}     & 64                   \\ \hline
\textbf{Optimizer}                       & Rmsprop              \\ \hline
\textbf{Learning rate}                   & \num{1e-5} decayed to \num{1e-7} \\ \hline
\textbf{Weight Decay}                    & \num{1e-4}                 \\ \hline
\textbf{Replay buffer size}              & \numprint{10000}                \\ \hline
\textbf{Target update every}             & \numprint{2000}                 \\ \hline
\textbf{Action classifier learning rate} & \num{1e-4}                 \\ \hline
\end{tabular}

\caption{Minigrid Network and traininig.}
\end{table}

\begin{table}[h]
\begin{tabular}{|l|l|}
\hline
\textbf{Word embedding size}             & 128                  \\ \hline
\textbf{Inventory RNN size}              & 256                  \\ \hline
\textbf{Description RNN size}            & 256                  \\ \hline
\textbf{Obs RNN size}                    & 256                  \\ \hline
\textbf{Fully Connected hidden size}     & 400                  \\ \hline
\textbf{Optimizer}                       & Rmsprop              \\ \hline
\textbf{Learning rate}                   & \num{1e-5} decayed to \num{1e-7} \\ \hline
\textbf{Weight Decay}                    & \num{1e-4}                 \\ \hline
\textbf{Replay buffer size}              & \numprint{10000}                \\ \hline
\textbf{Target update every}             & \numprint{2000}                 \\ \hline
\textbf{Action classifier learning rate} & \num{1e-4}                 \\ \hline
\end{tabular}

\caption{TextWorld Network and training.}
\end{table}

\end{document}